\documentclass[10pt,twocolumn,letterpaper]{article}

\usepackage[pagenumbers]{cvpr} 
\usepackage{graphicx}
\usepackage{caption}
\usepackage{listings}
\usepackage{dirtree}
\usepackage{multirow}
\usepackage{array}
\usepackage{tabularx} 
\usepackage{array}
\usepackage{subcaption}
\newcolumntype{C}[1]{>{\centering\arraybackslash}m{#1}}
\newcolumntype{L}[1]{>{\raggedright\arraybackslash}m{#1}}
\PassOptionsToPackage{dvipsnames,svgnames}{xcolor}
\usepackage{flowchart}
\usetikzlibrary{arrows}
\usepackage{tikz}
\usepackage{fontawesome5}
\usepackage[accsupp]{axessibility}

\lstdefinestyle{mystyle}{
    backgroundcolor=\color{backcolour},   
    commentstyle=\color{codegreen},
    keywordstyle=\color{magenta},
    numberstyle=\tiny\color{codegray},
    stringstyle=\color{codepurple},
    basicstyle=\ttfamily\footnotesize,
    breakatwhitespace=false,         
    breaklines=true,                 
    captionpos=b,                    
    keepspaces=true,                 
    numbers=left,                    
    numbersep=5pt,                  
    showspaces=false,                
    showstringspaces=false,
    showtabs=false,                  
    tabsize=2
}

\usepackage[dvipsnames]{xcolor}

\definecolor{cvprblue}{rgb}{0.21,0.49,0.74}
\usepackage[pagebackref,breaklinks,colorlinks,citecolor=cvprblue]{hyperref}

\def\paperID{15} 
\def\confName{CVPR}
\def\confYear{2024}

\title{Mitigating Challenges of the Space Environment for Onboard Artificial Intelligence: Design Overview of the Imaging Payload on  SpIRIT}

\author{Miguel Ortiz del Castillo\quad Jonathan Morgan \quad Jack McRobbie \quad Clint Therakam \quad Zaher Joukhadar \\ Robert Mearns \quad Simon Barraclough \quad Richard Sinnott \quad Andrew Woods \quad Chris Bayliss \\ Kris Ehinger \quad Ben Rubinstein \quad James Bailey \quad Airlie Chapman \quad Michele Trenti\\
The University of Melbourne\\
{\tt\small miguel.ortizdelcastillo@unimelb.edu.au}
}

\usepackage[T1]{fontenc}
\begin{document}
\maketitle

\begin{abstract}

Artificial intelligence (AI) and autonomous edge computing in space are emerging areas of interest to augment capabilities of nanosatellites, where modern sensors generate orders of magnitude more data than can typically be transmitted to mission control. Here, we present the hardware and software design of an onboard AI subsystem hosted on SpIRIT.  The system is optimised for on-board computer vision experiments based on visible light and long wave infrared cameras. This paper highlights the key design choices made to maximise the robustness of the system in harsh space conditions, and their motivation relative to key mission requirements, such as limited compute resources, resilience to cosmic radiation, extreme temperature variations, distribution shifts, and very low transmission bandwidths. The payload, called Loris, consists of six visible light cameras, three infrared cameras, a camera control board and a Graphics Processing Unit (GPU) system-on-module. Loris enables the execution of AI models with on-orbit fine-tuning as well as a next-generation image compression algorithm, including progressive coding. This innovative approach not only enhances the data processing capabilities of nanosatellites but also lays the groundwork for broader applications to remote sensing from space.

\end{abstract}    
\section{Introduction}
\label{sec:intro}

Nanosatellites represent a significant advancement in space technology, offering a compact and efficient alternative to traditional satellite designs for a growing range of applications. Their design, which often incorporates Commercial Off-The-Shelf (COTS) components, achieves considerable cost savings, thereby making space exploration more accessible, at the expense of higher risk. \cite{oberright2002nanosatellite}. However, the lower cost and rapid progress from conceptual design to flight naturally allows for a greater risk appetite, thus opening the opportunity to utilize modern compact sensors, even if the hardware is not explicitly rated for usage in space. Capable of performing a variety of tasks, these satellites facilitate a wide array of application-oriented opportunities, from detailed Earth and space observations to agile deployment of novel missions \cite{selva2012survey}, thereby offering promising prospects to transform the field.

Recent technological advancements in the miniaturization of computing modules have significantly expanded the capabilities of nanosatellite payloads. This breakthrough has facilitated the integration of diverse, computing devices within nanosatellites, empowering them to execute complex processing tasks previously thought to be impractical for such compact platforms \cite{furano2020towards}. This advancement in payload design opens up new avenues for on-board processing capabilities, enabling the transition of data processing tasks, once solely possible on ground systems, to direct execution in-the-loop while in orbit. A particularly promising opportunity is the deployment of sophisticated AI systems on these outer-edge devices, leveraging the enhanced computational power to enable near-real-time data analysis and decision-making directly on the satellite \cite{kothari2020final}.

Although COTS components on nanosatellites unlock significant potential, they also present distinct challenges for the successful deployment of AI systems on-board satellites. The operational environment for these AI solutions is marked by harsh conditions and a scarcity of resources, notably power (or often equivalently, ability to dissipate heat), memory,  computational capacity and communication bandwidth \cite{denby2019orbital}. This restricts both the complexity of the models and their adaptability through retraining and tuning once the satellite is in orbit. Furthermore, to operate effectively, the AI model requires robustness in the face of unexpected environmental shifts and data variations, as well as significant autonomy. The need for autonomy is accentuated by the communication challenges, characterized by constrained bit rates and limited opportunities for uplink and downlink of telemetry and telecommands. Moreover, the hardware design must be well executed to withstand the radiation environment specific to the target mission's orbit \cite{gutierrez2023toward} and to address the thermal challenges that arise from operating an edge device in the vacuum of space.

In response to the significant challenges identified in deploying AI systems on nanosatellites, the Melbourne Space Laboratory, in collaboration with national and international partners, has recently launched a nanosatellite. This platform hosts `Loris', a payload that integrates an NVIDIA Jetson Nano single-board computer with a multi-camera system, specifically engineered to facilitate AI operations in orbit. This endeavor aims to demonstrate the viability of sophisticated hardware solutions enabling on-board AI whilst addressing the complexities and challenges of the space environment. Our goals are to contribute to the evolving landscape of AI applications on nanosatellites, to support flexible and adaptable solutions, to enhance real-time data processing and to improve (in situ) orbital decision-making capabilities.

In this paper, we outline the design and functionality of our system, emphasizing the key design choices that have allowed us to address the challenges inherent in the space environment. 

\section{Related Work}
\label{sec:related_work}

Reduced launch costs have markedly simplified the process of deploying satellites into orbit, prompting researchers to delve into the potential of deploying COTS edge processors, with their capacity for low-latency and distributed computing, in space \cite{denby2019orbital}. A notable milestone in this domain is the European Space Agency's $\Phi$-Sat-1 mission, which pioneered the use of on-board AI for real-time data analysis \cite{giuffrida2021varphi}. Featuring an Intel Movidius Myriad 2 Visual Processing Unit (VPU), $\Phi$-Sat-1 manages to run deep Convolution Neural Network (CNN) models for precise cloud detection directly on the satellite, a testament to the potential of low-power COTS components in executing complex computations with less than 2W of power \cite{leon2021improving}. Furthermore, the Myriad 2 has undergone extensive evaluation for its suitability in space environments, showing resilience to radiation effects. The system experienced no Single Event Latch-up (SEL) at energies up to 8.8 MeVcm$^2$/mg, and displayed no evidence of sensitivity to cumulative Cobalt-60 radiation up to 49 krad in ground Total Ionizing Dose (TID) tests \cite{furano2020towards}, proving the viability of the Myriad 2 for space missions.

Building on the successful deployment of AI technologies in space missions, exemplified by $\Phi$-Sat-1 mission, the field is now rapidly progressing \cite{bayer2023reaching}. The emergence of compact devices equipped with powerful embedded Graphics Processing Units (GPUs), such as the ARM Mali and NVIDIA Jetson series, represents a significant advancement \cite{rad2023preliminary}. Notably, the NVIDIA Jetson stands out as a promising option with low-power, high capability and small form factor \cite{lofqvist2020accelerating}. Recognizing the benefits of incorporating such a module into a nanosatellite, several studies have assessed the Jetson Nano GPU's resilience in space-like environments. A recent experiment conducted at the U.S. Air Force Research Laboratory with Cobalt-60 irradiation indicates that the Jetson Nano could exceed a Total Ionizing Dose (TID) of 20 krad \cite{slater2020total}. This TID radation experiment indicated that the Jetson Nano TID provides acceptable risk levels for a nanosatellite mission lifetime ranging from 1.5 to 2 years in Low Earth orbits with approximately 3mm of aluminum shielding. Beyond evaluating the Jetson Nano's radiation endurance, research has also explored the radiation impact on micro-SD cards, a key component on some versions of these modules, revealing a tolerance of up to 40 krad TID \cite{duhoon2021total}. This combination of hardware resilience and strategic shielding underscores its capability to meet mission requirements despite the demanding conditions.

The foundation laid by studies such as  \cite{software_approach_ai_onboard}, \cite{zhou2021intelligent}, \cite{murbach2023brainstack}, and \cite{aerospace10020101} has significantly contributed to the initial understanding of on-board AI capabilities in space. However, a cohesive and practical approach to system design, specifically tailored to the harsh space environment, is still missing. In response, in this paper we present the design of the Loris payload and the mitigation strategies applied to provide a comprehensive framework for developing robust hardware and software AI systems suitable for space.

\section{Loris Payload Design}
\label{sec:sys_arch}

\begin{figure}[ht]
    \centering
    \includegraphics[width=0.85\linewidth]{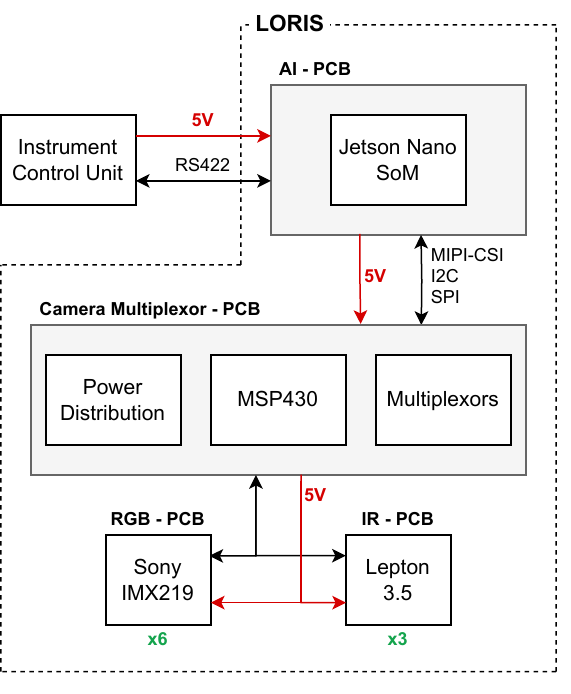}
    \caption{Loris payload architecture}
    \label{fig:architecture}

\end{figure}

The Loris payload, showcased in Figure \ref{fig:architecture}, consists of three primary elements: a Jetson Nano for processing, secured on the Loris carrier frame - a mechanical mount acting as a custom radiation shield and thermal sink, a camera multiplexer, and a collection of imaging sensors mounted on custom PCB adaptors. Central to the system is an NVIDIA Jetson Nano single-board computer, which enables on-board image processing and compression. This module interfaces with the Instrument Control Unit (ICU) responsible for coordinating all payload operations via a differential RS-422 full-duplex communication protocol. Through this connection, the AI board receives operational commands and transmits telemetry data, including image files for download.

To address the Jetson Nano's limitations, specifically its limited number of camera interfaces, a camera channel multiplexer has been designed. This module interfaces with up to 16 sensors, significantly expanding Loris's observational capabilities and versatility. By leveraging the Jetson Nano's native hardware support, the module enhances its capacity through the integration of additional channels using multiplexers. These are controlled by an MSP430 microcontroller, facilitating the accommodation of various types of sensing devices. The control of the cameras is managed through an I2C interface, offering two methods for receiving image data: either through eight 4-lane MIPI-CSI channels or eight SPI channels. For this mission, the payload's imaging array comprises six Sony IMX219 RGB cameras alongside three FLIR Lepton 3.5 long-wave infrared sensors, offering a diverse imaging variety, with sensors located on multiple sides of the spacecraft.

\begin{figure}[ht]
    \centering
    \includegraphics[width=1.0\linewidth]{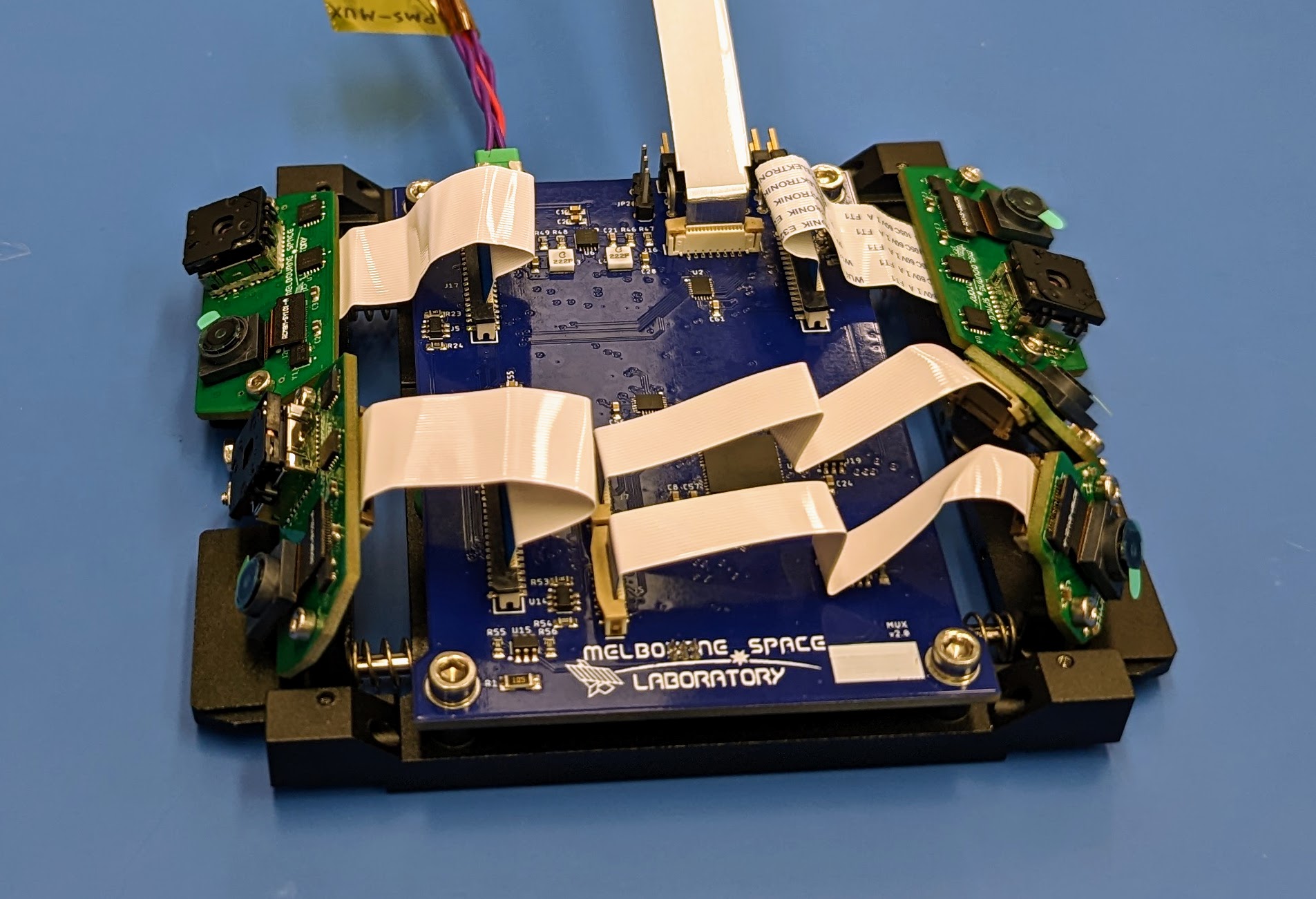}
    \caption{Loris Camera and Multiplexing Electronics Sub-module}
    \label{fig:mux}
\end{figure}

Summary specifications of the Jetson Nano selected for the mission are provided in Table \ref{tab:jetson}, while the image sensors' characteristics are outlined in Table \ref{tab:sensor-specs}. Figure \ref{fig:mux} illustrates the Loris Camera Multiplexing submodule, showcasing the integration of nine cameras and the associated multiplexing electronics, highlighting the payload's compact and efficient design, which includes a deployable arm to extend the cameras outside the nanosatellite once in orbit. 

\begin{table}[htb]
  \centering
  \begin{tabular}{@{}ll@{}}
    \toprule
    \textbf{Component} & \textbf{Specification} \\
    \midrule
    GPU & 128-core Maxwell \\
    CPU & Quad-core ARM A57 @ 1.43 GHz \\
    Memory & 4 GB 64-bit LPDDR4 25.6 GB/s \\
    Storage & On-board 16 GB eMMC \\
    \bottomrule
  \end{tabular}
  \caption{Specifications of the Jetson Nano on Loris.}
  \label{tab:jetson}
\end{table}

\vspace{-0.2cm}
The adoption of a multiplexing strategy for camera sensors in our payload mitigate the risk of individual sensor failures, enhancing the overall resilience of the imaging system. Employing cost-effective COTS sensors, such as the IMX219 and Lepton 3.5, aligns with the typical budget constraints of nanosatellite missions although this choice inherently carries a higher risk of component failure than space-grade cameras. By incorporating multiple sensors, our design not only mitigates this risk but also facilitates in-orbit performance validation of the sensors, drawing parallels to terrestrial testing conducted for the IMX219 \cite{Antonsanti2023}.

\begin{table}[htb]
  \centering
  \vspace{0.2cm}
  \label{tab:sensor-specs}
  \begin{tabular}{@{}L{3.4cm}C{1.9cm}C{1.9cm}@{}}
    \toprule
    \textbf{Parameter} & \textbf{Visible IMX219} & \textbf{IR Lepton 3.5} \\
    \midrule
    Resolution (px) & 3820 x 2464 & 160 x 120 \\
    FoV (deg) & 62.2 x 48.8 & 57 x 44 \\
    Pixel Size ($\mu$m) & 1.2 & 12 \\
    GSD@500 km (m/px) & 200 & 3,400 \\
    Frame-rate (fps) & 15$^*$ & 8.7 \\
    Spectrum ($\mu$m) & 0.4-0.7 & 8.0-14.0 \\
    Dynamic Range (°C) & N/A & -10–400 \\
    Exposure time range & 34$\mu$s-358ms & Auto \\
    \bottomrule
  \end{tabular}
  \begin{flushleft}
    \scriptsize{*Full resolution (frame rates vary based on resolution)}
  \end{flushleft}
  \vspace{-0.4cm}
  \caption{Loris Sensor Specifications}
  \label{tab:sensor-specs}
\end{table}

\section{Challenges and Mitigation Strategies}
\label{sec:sys_arch}

Deploying AI systems on nanosatellites poses unique challenges due to the extreme conditions of space, the inherently limited resources available on these compact platforms, and the continuously evolving demands characteristic of modern space missions. This section examines the principal challenges encountered during the conceptualization and development of the Loris payload, designed for a mission duration of two years within a 550 km Sun-Synchronous Orbit (SSO). These challenges include managing thermal constraints, ensuring radiation resilience, overcoming the limited communication bandwidth, optimizing scarce computational resources, and maintaining the in-orbit adaptability of AI algorithms.

\begin{figure*}[t]
    \centering
    \begin{subfigure}[b]{.48\textwidth}
        \includegraphics[width=\textwidth]{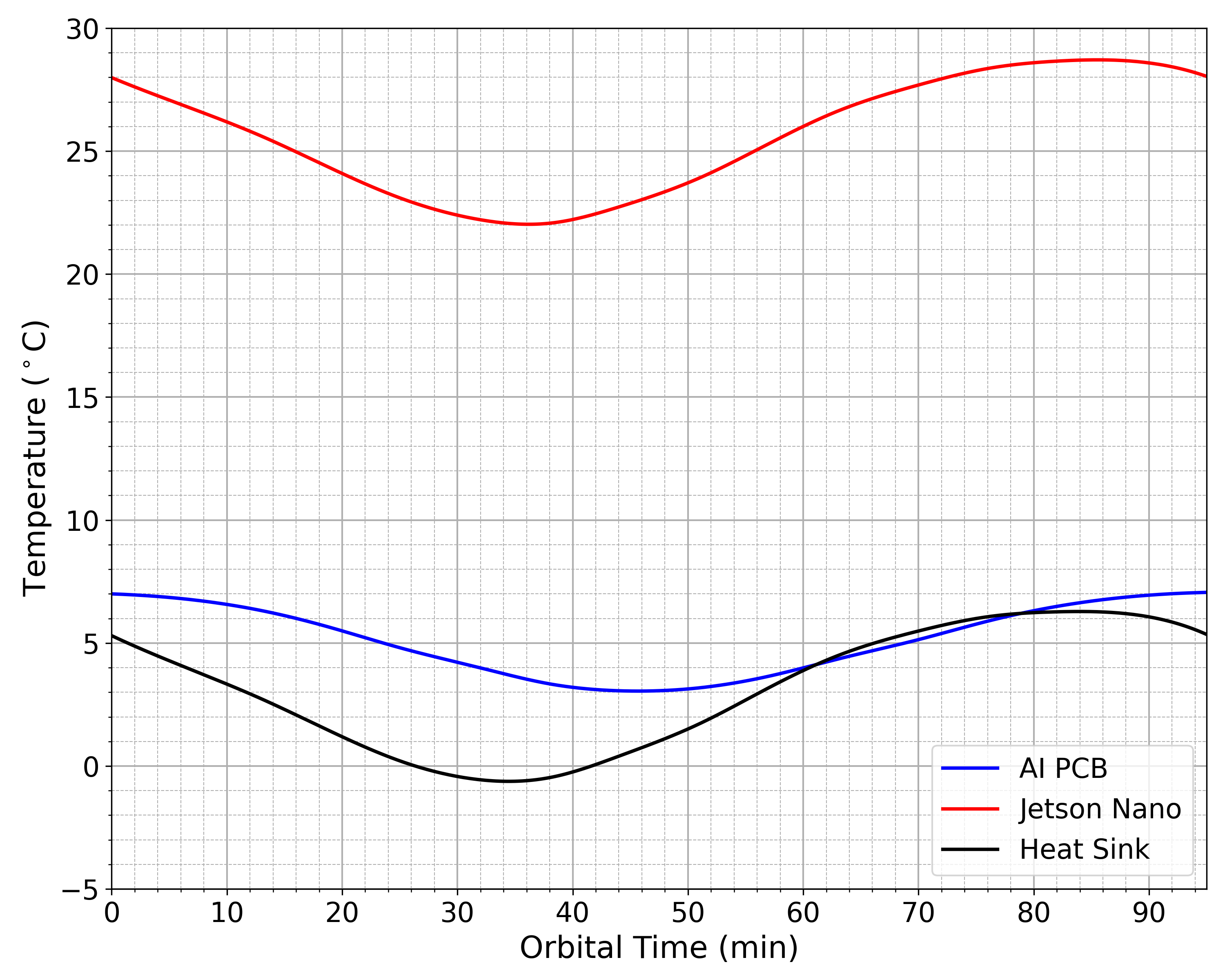}
        \caption{Loris payload active}
        \label{fig:active}
    \end{subfigure}
    \hfill
    \begin{subfigure}[b]{.48\textwidth}
        \includegraphics[width=\textwidth]{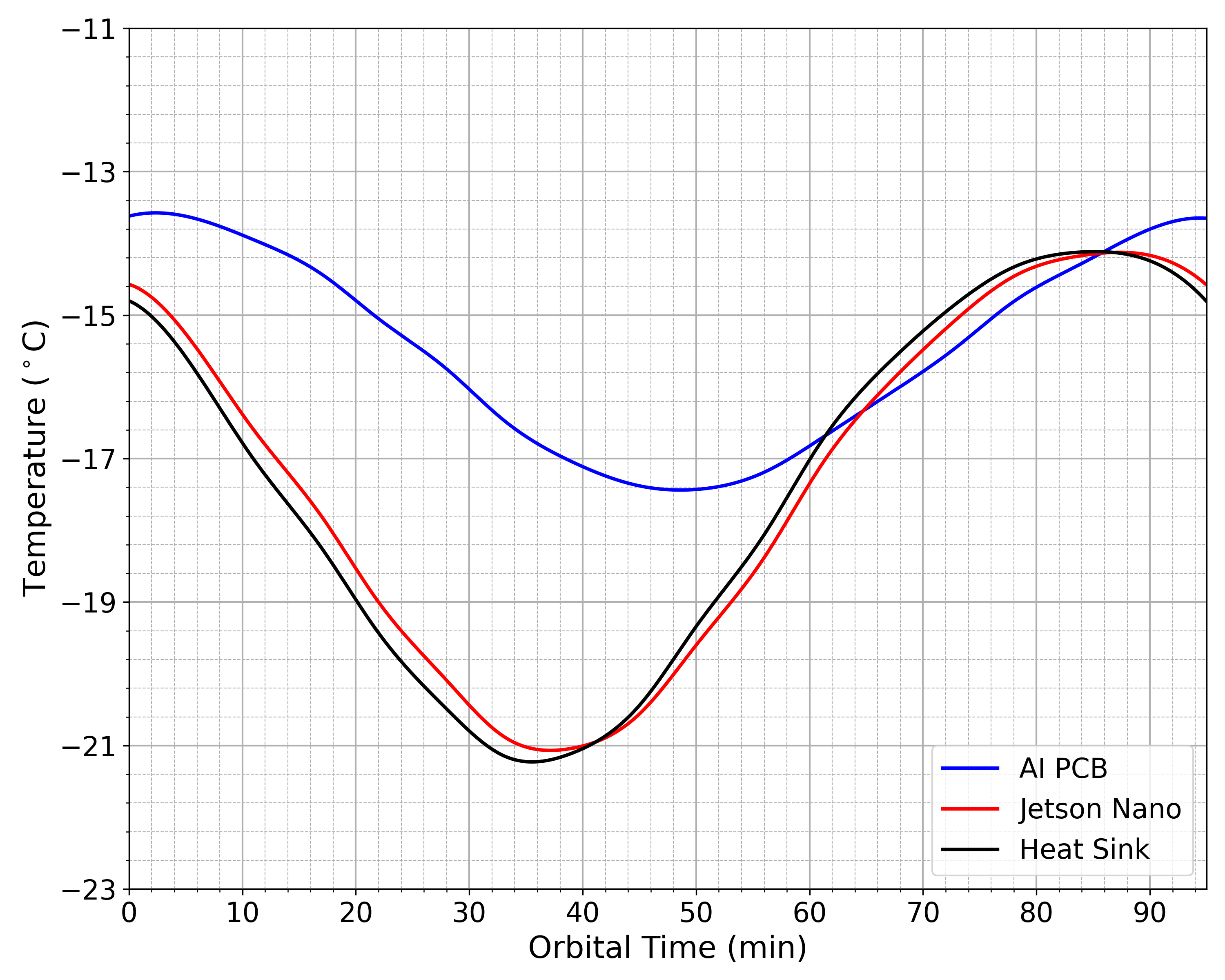}
        \caption{Loris payload idle/inactive}
        \label{fig:idle}
    \end{subfigure}
    \caption{Results from worst case thermal simulation of Loris on board the host mission (Polar Sun Synchronous Orbit at approx. 500km).}
    \label{fig:thermalSimulation}
\end{figure*}

\subsection{Thermal Management}

Considering the Jetson Nano's power consumption, estimated at 5W,
appropriate thermal management and design are important to prevent both overheating when running, and overcooling when idle or off. Furthermore, on-orbit thermal management strategies must accommodate fluctuating thermal environmental loads encountered during a single orbit (eclipse versus direct sunlight). In vacuum, traditional convectional cooling strategies, such as finned heat-sinks, are ineffective, necessitating only conductive and radiative thermal heat transfer.

Within the Loris payload, the Jetson Nano's CPU, GPU, and critically, the power management chip, are all conductively coupled to the Loris carrier frame, a singular aluminium frame designed to closely follow the contours of the module. The frame here acts as both a thermal sink for the board, increasing the thermal inertia of the processors, thereby reducing the extremes of the Nano's transient thermal range as the spacecraft enters and exits eclipse, and as a thermally conductive path from the Loris payload to the spacecraft heat rejection mechanisms. Sufficient thermal contact between the Jetson Nano and the Loris carrier frame is ensured via the use of a compliant, space-rated thermal filler (Laird Tflex). This filler material is conceptually similar to traditional thermal paste used between a processor and finned heat-sink, but is suitable for in-orbit use.

To validate Loris thermal performance, thermal simulations were conducted to assess the Jetson Nano's temperature extremes using ESATAN-TMS, an industry-leading space thermal simulation software. It can be used to model a thermal system with lumped-mass thermal objects (called nodes), and iteratively solves the Temperature-Time ODEs at each orbital time-step. These simulations produce both steady-state and transient results under assumptions about environmental thermal loads, and operational conditions of the subject (power consumption/dissipation). A thermal model encompassing the entire nanosatellite was developed for the mission, comprising 200 nodes, of which only three are attributed to the Loris payload. Nevertheless, the fidelity of this model, is deemed sufficiently high to assess the temperature range of Loris.

The results in Figure \ref{fig:thermalSimulation}, show that Loris will experience a sinusoidal thermal profile while in-orbit due to the spacecraft transitioning in and out of eclipse, and a phase lag of the carrier PCB, while the Nano moves in sync with the thermal sink. Due to the driving requirements of the primary payload, SpIRIT has been designed to err on the cold side of the temperature limits of other payloads (the Jetson Nano included), as can be seen in the idle/inactive case of Figure \ref{fig:thermalSimulation}. The Jetson Nano remains fluctuating between -14 to -21 $^\circ$C in its inactive state, during the typical cold rest state of the nanosatellite. Nevertheless, this is above both its non-operational ($T>-40$$^\circ$C), and operational cold limits ($T>-25$$^\circ$C). Thus, the Jetson Nano can safely be powered on at any-time in orbit. 

\begin{figure*}[t]
    \centering
    \includegraphics[width=\textwidth, height=7cm]{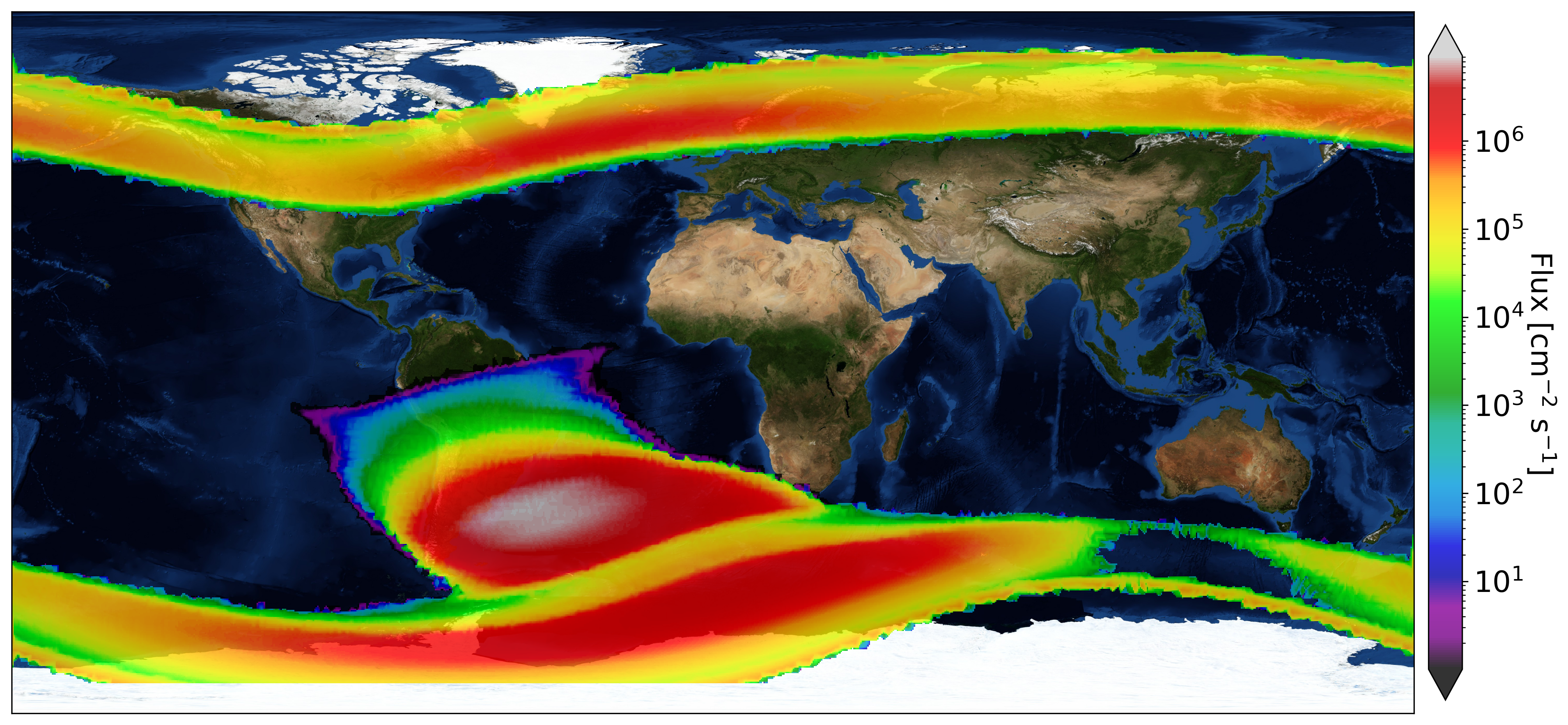}
    \caption{Radiation environment for proton and electron fluxes for Loris}
    \label{fig:RadiationEnv}
\end{figure*}

During typical full Loris operations, the simulations predict the Jetson Nano will reach a maximum temperature of +22 to +29 $^\circ$C over the course of an orbit, with a difference of approximately 43$^\circ$C between the maximums of the active and inactive cases. The maximum predicted temperatures are well below the Jetson operational maximum of $T<+97$$^\circ$C, with a margin of 68 $^\circ$C. The simulated positive margin to both the operational maximum and to the non-operational minimum demonstrates complete operational capability from a thermal perspective, of the Nano while in orbit. A possible alternative strategy, in the event that the Loris payload is consistently reaching its maximum allowable temperatures, is to duty-cycle the system, and therefore reduce the heat dissipation of the Jetson Nano; however, it is not anticipated that the onboard Jetson Nano will require duty cycling as the system has been tuned to the cold end of the allowable temperature envelope.

In addition to the above simulations, Loris has undergone system-level thermal testing as part of the nanosatellite environmental test campaign. During this campaign, the spacecraft, and therefore Loris, underwent various thermal tests in a simulated space environment. During this test campaign, Loris was subjected to multiple hot and cold cycle extremes in vacuum at predicted system level maximum and minimum temperature extremes. At these extremes, short functional test checkouts were conducted to ensure the payload was behaving nominally - with Loris passing all checkouts. The spacecraft also underwent thermal balance testing wherein both power dissipation in the spacecraft and the environment are kept constant, allowing the spacecraft and all its payloads to reach thermal equilibrium. This steady-state behaviour allows thermal models used for simulation to be correlated with real-world results, increasing the confidence in orbital thermal predictions. These thermal balance tests included running Loris continuously for 10 hours. During this, Loris performed admirably, reaching a maximum temperature of +43 $^\circ$C. It should be noted, that this differs from the predicted orbital maximum by $\sim$20$^\circ$C, primarily as a result of the different environmental conditions. The Thermal balance test lasted for $\sim$10 hours before reaching steady-state, while a single orbit of the Loris payload will be only 90 mins; insufficient time to reach thermal equilibrium before the external conditions change.

From the thermal simulations and physical testing the Loris payload and its Jetson Nano have undergone, it is evident that the thermal management strategies employed will be effective in maintaining the Jetson Nano's temperature within its operational and non-operational limits.

\subsection{Radiation Environment}

Space missions face the challenge of ionizing radiation, which threatens the functionality of satellite electronics and data integrity. The LEO radiation environment is characterized by the presence of both high-energy protons and lower-energy electrons that can cause critical damage to the spacecraft operational systems \cite{gutierrez2023toward}. Therefore, understanding the orbit's radiation environment is crucial for safeguarding Loris functionality and mission success.

The integral flux of protons and electrons along the orbital path of Loris was simulated using the software SPENVIS \cite{lawrence_space_2010} and the European Cooperation for Space Standardization (ECSS-E-ST-10-04) standards. This method provides an overview of the radiation environment expected during the orbit. Specifically, Figure \ref{fig:RadiationEnv} integrates the data for proton flux (greater than 0.1 MeV) from NASA's AP-8 MIN model and electron flux (greater than 0.04 MeV) from the NASA AE-8 MAX model. As high-energy particle flux is directly correlated with the risk of Single Event Effects (SEE), the resulting radiation map highlights regions where the risk of SEEs, such as bit flips in volatile or non-volatile memory, is increased. Given its operation in a sun-synchronous orbit, Loris will be exposed to the intensified radiation at the poles and within the South Atlantic Anomaly, as compared to an equatorial orbit, increasing the need for a resilient software architecture to address transient effects which may occur while traversing those regions.

\begin{figure}[ht]
    \centering
    \includegraphics[width=0.80\linewidth]{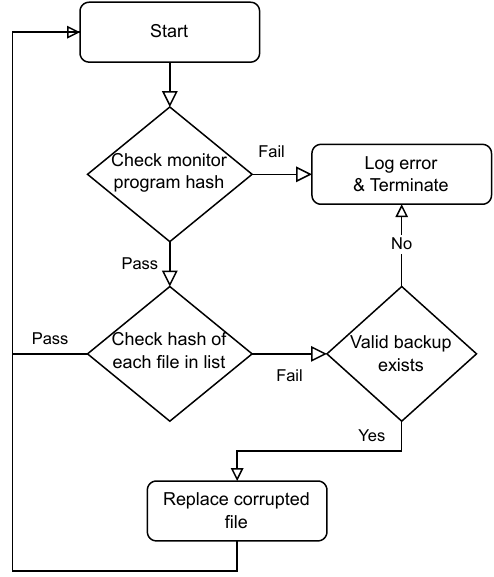}
    \caption{Loris Software Integrity Monitoring Architecture}
    \label{fig:monitor}
\end{figure}

\begin{figure}[b]
    \centering
    \includegraphics[width=1.0\linewidth]{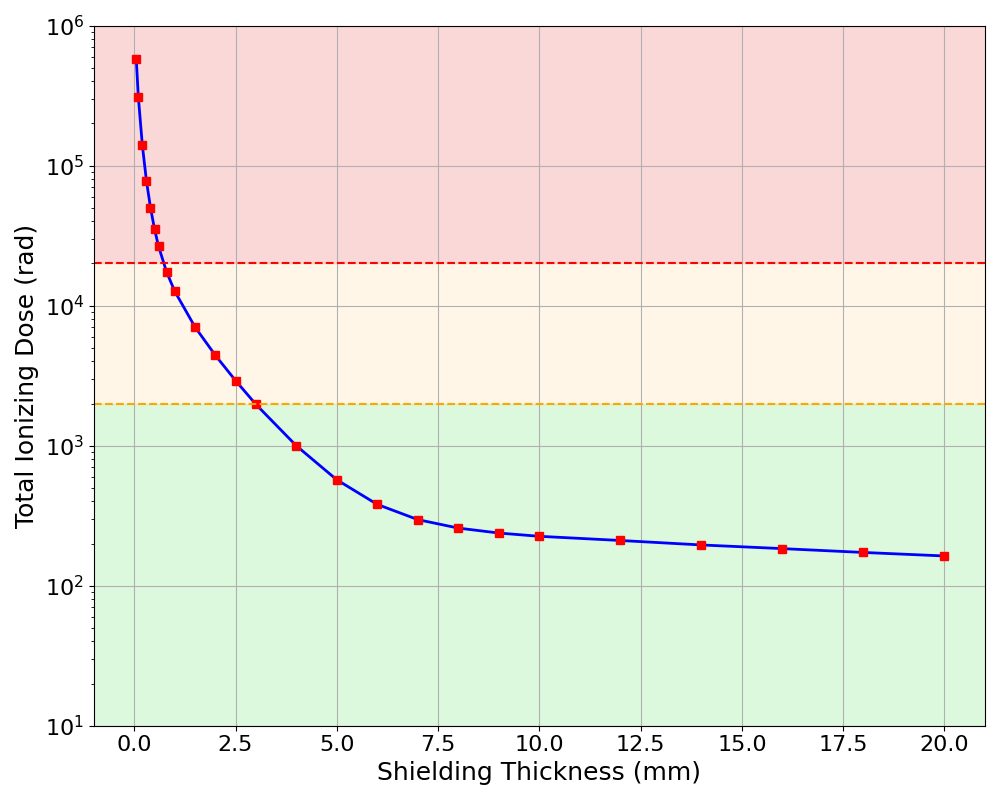}
    \caption{Shielding vs TID for Loris calculated for 2 year mission}
    \label{fig:TID}
\end{figure}

Loris mitigates SEE that could corrupt its core code and important internal data by running a dedicated subroutine for error detection and correction. This approach not only enhances system reliability but also significantly extends the operational lifespan of the payload by reducing the likelihood that critical functionalities will be compromised. The monitoring routine maintains a list of files and their MD5 hash. Periodically, it performs a MD5 hash calculation for both backup and live versions of the files on this list and compares the outcome to the stored hash values. If the file is found to be corrupted it is replaced with a backup, provided the backup is not also corrupted. The program does not distinguish between files in use and backups, so if a backup is corrupted it will be checked and restored from the live file. Important files can have increased redundancy by storing multiple  backups, where each backup points to a different copy. Finally, if all copies of a file are corrupted the program logs the error so a valid copy of the file can be re-uploaded to the spacecraft.  The control flow is shown in Figure \ref{fig:monitor}. The monitoring software runs every 60 seconds, in an experiment on a similar device, the Jetson Xavier NX, less than $1 \times 10^{-2}$ SEEs were expected per orbit in low Earth orbit\cite{Rodriguez_2022}, so this is a conservative interval.  In addition to active integrity and consistency checks, Loris adopts operational strategies to minimize SEE exposure in volatile memory by refraining from activating the payload while traversing regions known for heightened radiation flux, specifically the polar areas and the South Atlantic Anomaly, as illustrated in Figure \ref{fig:RadiationEnv}.

Following the identification of critical radiation zones via SPENVIS simulations, the SHILDOSE-2 model \cite{seltzer1994updated} has been utilized to estimate the Total Ionizing Dose (TID), guiding the optimization of shielding thickness within the stringent constraints of the mission's mass budget. This optimisation seeks to maximize radiation protection with minimal mass increase. Figure \ref{fig:TID} delineates the TID for a given shielding thickness, distinguishing three distinct regions: the red zone indicates expected doses exceeding 20 krad, the yellow zone represents doses between 20 krad and 2 krad, and the green zone signifies doses below 2 krad. A derating factor of x10 over thefindings presented in \cite{slater2020total}, in which 20krad is provided as a safe upper limit for the Jetson Nano, has been adopted to ensure ample margin and safeguard against system degradation. Consequently, a shielding thickness of 3mm was identified as necessary to achieve a TID of 2 krad throughout the mission's lifespan (see Figure~\ref{fig:TID}). The requisite shielding has been incorporated into the design of the Loris carrier frame.

\subsection{Limited Bandwidth}
\label{sec:compression}

\begin{figure*}
  \centering
  \begin{subfigure}{0.33\textwidth} 
    \centering
    \includegraphics[width=0.9\linewidth, height=4.25cm]{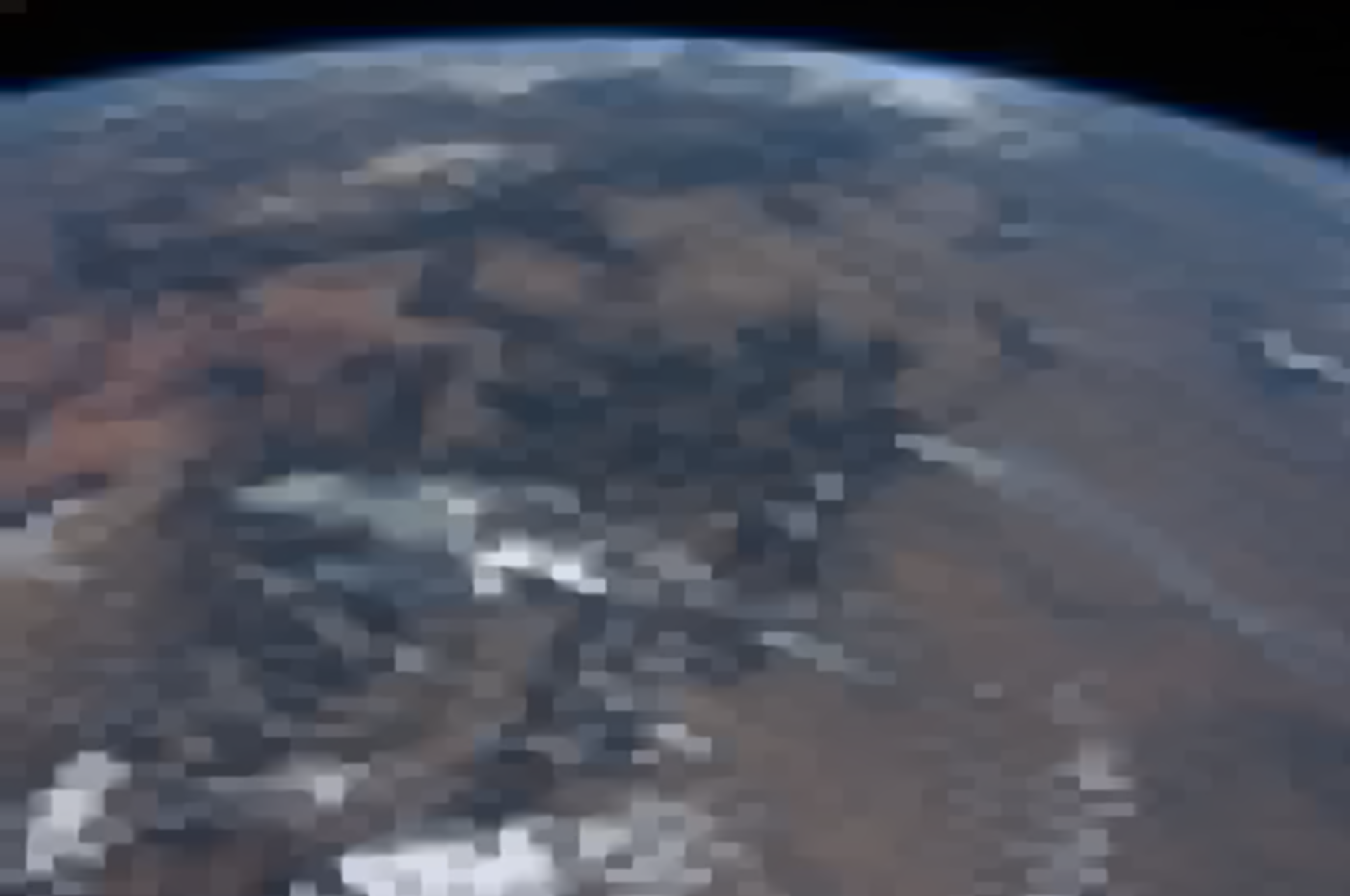}
    \caption{8kb - 1\% Preview}
    \label{fig:8kb}
  \end{subfigure}%
  \begin{subfigure}{0.33\textwidth}
    \centering
    \includegraphics[width=0.9\linewidth, height=4.25cm]{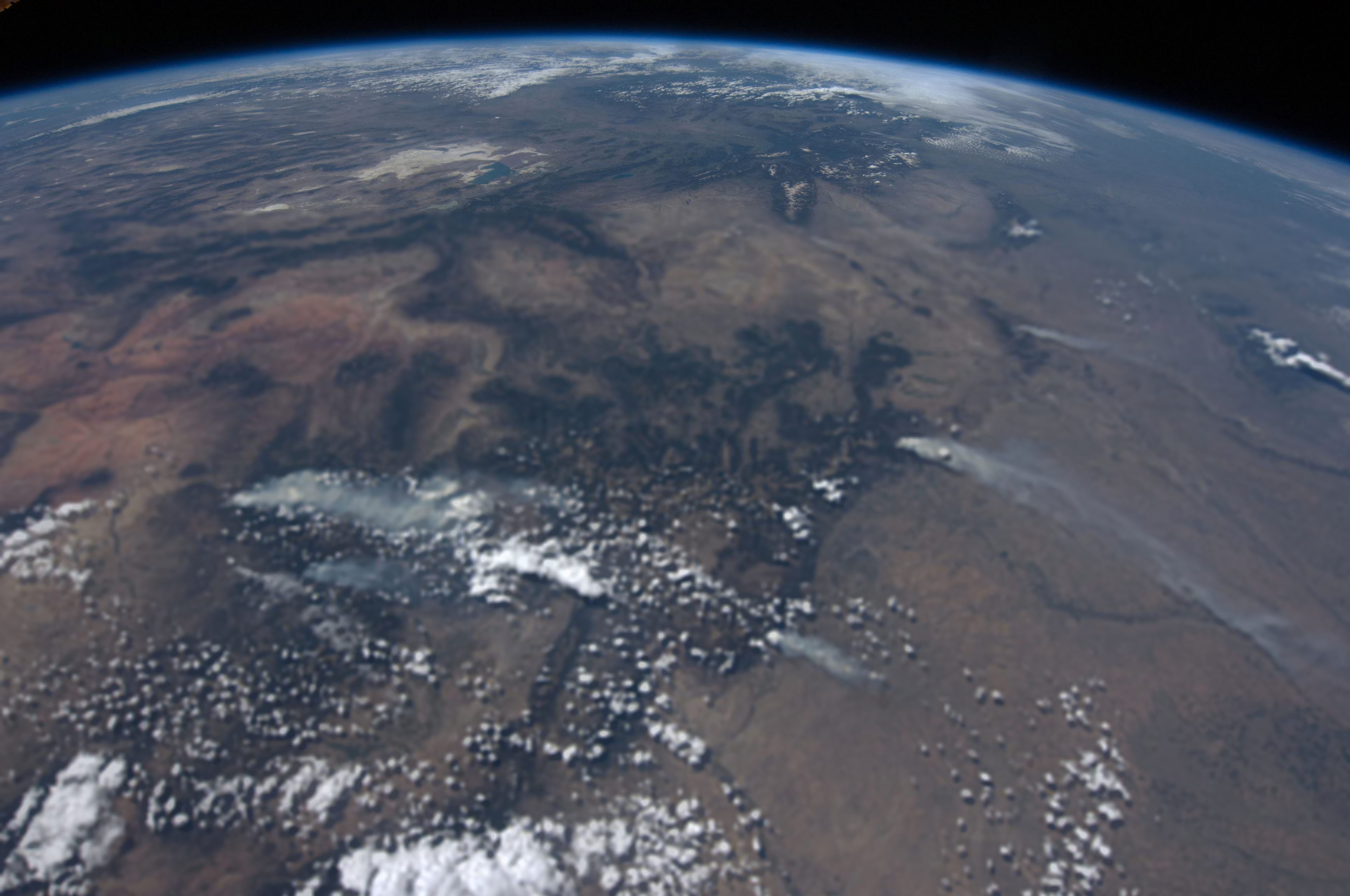}
    \caption{250kb - 30\% Preview}
    \label{fig:250kb}
  \end{subfigure}%
  \begin{subfigure}{0.33\textwidth}
    \centering
    \includegraphics[width=0.9\linewidth, height=4.25cm]{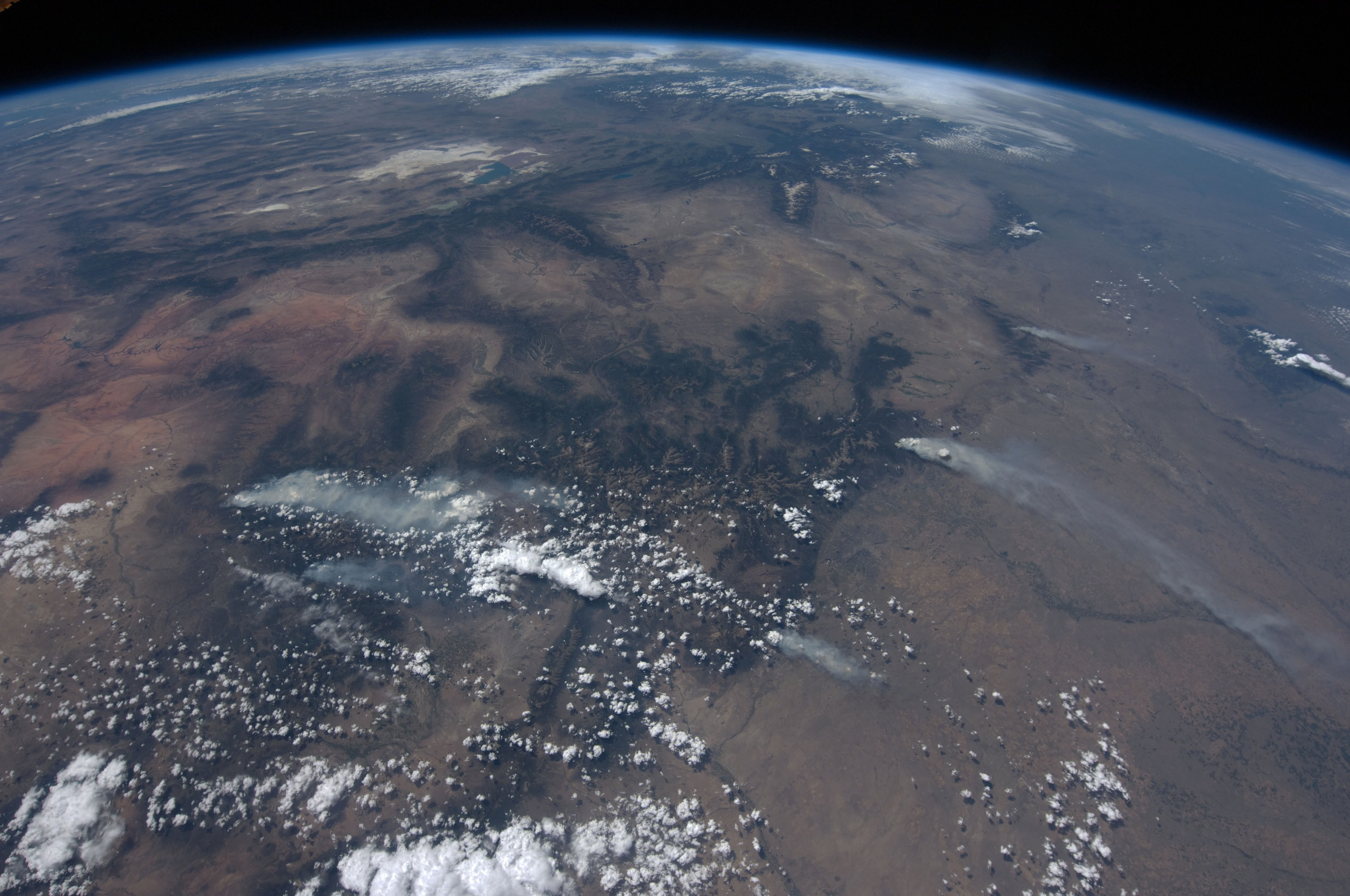}
    \caption{758kb - Full Jpeg-XL File}
    \label{fig:758kb}
  \end{subfigure}
  \caption{Progressive Coding Example - Using reference image from \cite{NASA_ISS031_E_148504_2012} Base Image: 1600kb}
  \label{fig:progressive_decoding}
\end{figure*}

One of the foremost challenges in space operations, particularly with nanosatellites, is the management of transmission bandwidth. The specific nature of these missions often limits their data transmission and reception capabilities, to communication with one or a few ground stations with a handful of short transmission windows a day (depending on orbit and ground station location). Even when data is downlinked through S-band for a typical bandwidth of ~2Mbits per second rather than UHF (9.600 kbits), modern cameras generate more data than can be retrieved, when not run at an extremely lowered duty-cycle. For SpIRIT, the ground segment consists of one UHF ground station for duplex communication with four available windows with the satellite each day, each lasting approximately 10 minutes, and one S-band ground station with simplex (satellite to ground) capabilities. This operational constraint significantly limits bandwidth, with our estimated uplink capacity set between 100KB and 200KB per day, and downlink capacity for Loris capped at 1MB per day after accounting for data downlink requirements from other payloads. To contextualize these figures, an uncompressed image captured by the satellite's camera has a resolution of 3820 x 2464 pixels in the RGB format, exceeding the daily data cap by a factor of 28 for a 24 bit RGB palette. The volume of data generated by just a single image of this quality highlights the bandwidth challenge, and underlines the critical need for efficient data management strategies, including advanced compression techniques and selective transmission. 

To address this bandwidth challenge, Loris implements the JPEG-XL algorithm for efficient on-orbit compression of image data captured by both RGB and IR sensors.  Choice of image compression algorithms is always a trade off between compression ratio, speed, and power-usage \cite{Alves}. Past image compression applications for nanosatellites have focused on optimising for low power usage \cite{Sendamarai_Giri} however due to the processing power available in the Jetson Nano for this mission, Loris compression aims to prioritise bandwidth limitations instead. Our aim is to test in orbit the use of the JPEG-XL compression for space applications due to its high compression ratio at `visually lossless' settings and the progressive coding feature \cite{Alakuijala2019}. JPEG-XL is an advanced image coding system based on a novel colour-space transform, variable sized discrete cosine transform and adaptive quantization and prediction steps \cite{Alakuijala2019}. To the best of our knowledge Loris' implementation is the first system that executes this algorithm in-orbit for space applications. 

\subsubsection{Lossless vs Lossy Compression}

When choosing a compression algorithm the first factor to consider is lossless or lossy compression. In lossless compression the original data can be perfectly reconstructed from the compressed data, whereas for lossy compression, the uncompressed data differs from the original, such that some information is lost. As Loris' cameras are not designed for high-fidelity scientific data collection but rather for technology, scientific demonstration, and public outreach, perfect reconstruction is not necessary. Thus, we use lossy compression for all images to achieve higher compression ratios. An advantage of JPEG-XL is that it can perform both lossy and lossless compression, so combined with Loris's re-configurable software architecture it is still possible to use lossless compression on an as-needed basis. 

\subsubsection{Progressive Coding}

Progressive coding encodes the compressed image bit-stream such that it can be decoded from partial transmission. As more of the image data is received the quality of the decoded image increases. This feature is particularly useful for nanosatellites as it allows efficient use of the limited bandwidth, as image transmissions can be stopped once the downlinked image is of sufficient quality \cite{Zhang2018}. JPEG-XL allows the use of progressive coding including placing the DC components of the discrete-cosine transform at the start of the bit-stream, allowing for decoding of the image with as little as 1\% of the image data, demonstrated in Figure \ref{fig:progressive_decoding}. During early in-orbit operations, this Loris feature has demonstrated its value by allowing evaluation of image acquisition using thumbnails of just 2kB in size, allowing tens of thumbnails to be downloaded per day.  

\subsubsection{Comparison with CCSDS Standard}

The Consultative Committee for Space Data Systems (CCSDS) produced an image data compression standard in 2017 that describes the recommended specifications for two-dimensional spatial data from payload instruments \cite{book_ccsds}. The CCSDS standard is based on the JPEG2000 format with reduced complexity to enable easier implementation on limited spacecraft hardware \cite{book_ccsds_2}. A key difference between JPEG-XL and the CCSDS standard is the colour space - the CCSDS standard supports only a single colour channel (i.e. grayscale images), whereas JPEG-XL uses a novel colour space, XYB, modelled on the human visual system \cite{book_ccsds_2} \cite{Alakuijala2019}. The mapping of the RGB colour space to XYB allows efficient encoding of redundant data between the colour channels, whereas the CCSDS standard would require encoding of each channel separately. 
As JPEG-XL targets a human visual system it is ideally suited for the RGB inspection cameras used by Loris, while the single channel approach of CCSDS is more suited to scientific instruments that may have a single channel \cite{jpegXL_benchmark}. The benefits of the CCSDS Standard, namely implementation ease, low-algorithmic complexity and memory efficiency, are not required on general purpose hardware such as the Jetson Nano.

\subsection{Computational Resources Limitations}

The Nvidia Jetson Nano equipped with a 128-core Maxwell GPU hosted on Loris does not match the computational power of terrestrial AI GPUs such as the A100 or H100. Yet, the inclusion of the Maxwell GPU represents a significant advancement in space computing technology, transitioning from predominantly CPU-based computations to more parallelizable GPU-enabled processing. Still, we note that the Jetson Nano model comes with its limitations, specifically, 4GB of RAM and 16GB of embedded Multi-Media Card (eMMC) storage, table \ref{tab:jetson}. Furthermore, the Linux operating system and housekeeping functions consume a significant portion of this storage, effectively leaving approximately 2GB free for operational use. Consequently, while on-board data processing benefits from enhanced computational capabilities, it still faces substantial constraints in terms of available storage space relative to the uncompressed size of an image collection from the cameras.

These limitations significantly influence the range of machine-learning models deployable on this hardware framework. For instance, lightweight models such as MobileNet can be used for real-time inference due to their efficient architecture and lower computational demands. MobileNet, specifically designed for mobile and edge devices, requires approximately 569 million operations for a single forward pass and can fit within the memory constraints of the Jetson Nano \cite{kristiani2020optimization}. In contrast, larger models with billions of parameters, common in advanced AI applications, exceed the Jetson Nano's processing and storage capabilities, necessitating a careful selection of algorithms that strike a balance between performance and resource consumption. This limitation necessitates innovative approaches to model optimization and compression to leverage the available computational resources effectively while navigating the storage constraints inherent to nanosatellites.

\subsection{In Orbit Adaptability}

As described above, one of the most significant constraints imposed on an imaging system on board a nanosatellite is management of transmission bandwidth. Therefore, accurate labeling of archived on-board data is extremely valuable to efficiently allocate downlink resources. As a proof of concept, Loris implements a cloud detection model and labels images autonomously, thereby providing an indication of an image's utility (whether it contains clouds or not).

One challenge with this approach is model fine-tuning and retraining.  Loris leverages an innovative approach known as the Ground Truth (GT) Factory, situated at the ground station. During its mission, Loris will gather metadata for each captured image, such as UTC time and GPS coordinates. Since direct image download is not feasible, the GT Factory utilizes the GPS data and time stamps from the captured images to infer estimated labels through correlation with data from alternative satellite sources or ground-based observations. These inferred labels, processed on the ground, are then relayed back to the satellite. This process facilitates dynamic, in-orbit refinement of the AI model, exemplifying a method for enhancing AI model accuracy through remote fine-tuning.

\section{Conclusion and Future Work}
\label{sec:conclusion}

In this paper, we presented the Loris imaging and onboard AI payload that was successfully launched on SpIRIT in December 2023. We examined the challenges of acquiring and processing images on a nanosatellite, particularly focusing on radiation and thermal management issues, limited downlink bandwidth, storage constraints for AI models, and innovative approaches for labeling images in orbit using metadata provided by the satellite. By implementing targeted solutions to these critical challenges, our payload design addresses the operational complexities associated with deploying AI in space and presents a framework for future missions to enhance AI deployment in space environments efficiently.

The payload, alongside the satellite platform, is currently in the commissioning phase on-orbit. To date, all sensors have successfully undergone initial checkouts, and preliminary images have been successfully downloaded, indicating effective operation and demonstrating the first use of JPEG-XL in orbit. Additionally, the Jetson Nano has demonstrated great performance, reinforcing its capability for advanced processing in space. These initial accomplishments provide a strong basis for future in-depth testing and operational endeavors, signaling a positive trajectory for the mission's progress.

\section*{Acknowledgment}
This research was partially funded by the Office of National Intelligence - National Intelligence and Security Discovery Research Grant program for NI220100072 and funded by the Australian Government.
{
    \small
    \bibliographystyle{ieeenat_fullname}
    \bibliography{main}
}

\end{document}